\documentclass[conference]{IEEEtran}
\IEEEoverridecommandlockouts
\usepackage{cite}
\usepackage{amsmath,amssymb,amsfonts}
\usepackage{algorithm}
\usepackage[noend]{algpseudocode}
\usepackage{graphicx}
\usepackage[lofdepth,lotdepth]{subfig}
\usepackage[font=footnotesize,labelfont=bf]{caption}
\usepackage{textcomp}
\usepackage{xcolor}
\usepackage{float}
\usepackage{enumitem}
\usepackage{booktabs}
\usepackage{multirow}
\usepackage{array}
\usepackage{authblk}
\usepackage[square,numbers]{natbib}
\usepackage{microtype}      
\def\BibTeX{{\rm B\kern-.05em{\sc i\kern-.025em b}\kern-.08em
    T\kern-.1667em\lower.7ex\hbox{E}\kern-.125emX}}

\makeatletter
\newcommand*{\inlineequation}[2][]{%
  \begingroup
    \refstepcounter{equation}%
    \ifx\\#1\\%
    \else
      \label{#1}%
    \fi
    \relpenalty=10000 %
    \binoppenalty=10000 %
    \ensuremath{%
      #2%
    }%
    ~\@eqnnum
  \endgroup
}
\makeatother

\makeatletter
\newcommand{\printfnsymbol}[1]{%
    \textsuperscript{\@fnsymbol{#1}}%
}
\makeatother

\usepackage{fancyhdr,lipsum}
\fancypagestyle{firstpage}{
  \fancyhf{}
  \fancyhead[C]{To appear at the 57th Design Automation Conference (DAC), July 2020, San Francisco, CA, USA.}
  \fancyfoot[C]{\thepage}
}

\begin{document}
\algnewcommand{\LineComment}[1]{\State \(\triangleright\) #1}

\title{Q-CapsNets: A Specialized Framework for Quantizing Capsule Networks\\
\vspace*{-4mm}
}

\author[1]{Alberto Marchisio\printfnsymbol{1}\thanks{\printfnsymbol{1}These authors contributed equally}}
\author[2]{Beatrice Bussolino\printfnsymbol{1}}
\author[1]{Alessio Colucci}
\author[2]{Maurizio Martina}
\author[2]{Guido Masera}
\author[1]{Muhammad Shafique \vspace*{-3mm} }
\affil[ ]{\textsuperscript{1}Technische Universit\"{a}t Wien (TU Wien), Vienna, Austria, \hspace{0.2cm} \textsuperscript{2}Politecnico di Torino (PoliTo), Turin, Italy}
\affil[ ]{Email: \{alberto.marchisio, alessio.colucci, muhammad.shafique\}@tuwien.ac.at}
\affil[ ]{\{beatrice.bussolino, maurizio.martina, guido.masera\}@polito.it \vspace*{-5mm}}

\renewcommand\Authsep{, }
\renewcommand\Authands{ and }

\maketitle
\thispagestyle{firstpage}

\begin{small}
\begin{abstract}
Capsule Networks (CapsNets), recently proposed by the Google Brain team, have superior learning capabilities in machine learning tasks, like image classification, compared to the traditional CNNs. However, CapsNets require extremely intense computations and are difficult to be deployed in their original form at the resource-constrained edge devices.
This paper makes the first attempt to quantize CapsNet models, to enable their efficient edge implementations, by developing a specialized quantization framework for CapsNets. We evaluate our framework for several benchmarks. On a deep CapsNet model for the CIFAR10 dataset, the framework reduces the memory footprint by 6.2x, with only 0.15\% accuracy loss. We will open-source our framework at https://git.io/JvDIF in August 2020.
\end{abstract}

\begin{IEEEkeywords}
Capsule Networks, Quantization, Compression.
\end{IEEEkeywords}
\vspace*{-1mm}
\end{small}


\section{Introduction}
\vspace*{-1mm}
Since the introduction of AlexNet~\cite{krizhevsky2012} in 2012, the interest in deep neural networks (DNNs) has grown steadily. Many models have been subsequently developed, achieving good accuracy in different tasks, such as object detection, computer vision and natural language processing. To achieve high accuracy, very deep and large DNNs models have been developed, for instance, from LeNet-5~\cite{lecun1998} having five layers to ResNet-152~\cite{he2016} having 152. Consequently, DNNs have a huge number of parameters, i.e., weights and biases, and their deployment in IoT systems is a challenge in terms of memory and computational resources. For example, the AlexNet requires 250MB memory for 60M parameters stored as 32-bits \verb|float|. These memory and computational requirements make DNNs unsuited for mobile and embedded devices. Much effort has been dedicated towards \textit{compressing DNN models} to address this problem. 
Quantization allows to significantly reduce the DNN model size, as well as enabling their applicability on different computing platforms like GPUs and FPGAs. In the literature, several quantization methods have been proposed~\cite{courbariaux2015}\cite{hubara2016}\cite{han2016}\cite{gysel2016}\cite{Lin2016}\cite{Vanhoucke2011}\cite{Jacob2017}\cite{anwar2015}\cite{sakr2019}. 

\begin{figure}[h]
    \centering
    \vspace*{-3mm}
    \includegraphics[width=0.9\linewidth]{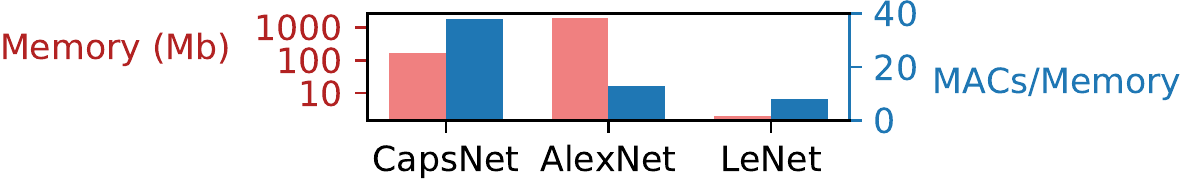}
    \vspace*{-1.5mm}
    \caption{\textit{Memory requirements and (Multiply-and-Accumulate operations vs. memory) ratio (MACs/memory) for ShallowCaps~\cite{sabour2017}, AlexNet~\cite{krizhevsky2012} and LeNet~\cite{lecun1998}.}}
    \label{fig:mac_memory}
    \vspace*{-2mm}
\end{figure}

Meanwhile, to improve the learning capabilities and accuracies of DNNs, the researchers at Google~\cite{sabour2017} introduced a novel DNN structure called \textit{Capsule Network (CapsNet)}, where individual neurons are substituted with \textit{capsules}, i.e., vectors of neurons. To overcome the loss of information introduced by the pooling layers, the pooling is substituted by a \textit{dynamic routing} process between the capsules of adjacent layers. As a drawback, CapsNets are much more challenging in terms of their memory requirement, memory bandwidth and energy consumption for the computational resources, compared to the traditional DNNs. To demonstrate this fact, we compare the CapsNet architecture introduced in~\cite{sabour2017}\footnote{We will refer to the CapsNet architecture introduced in~\cite{sabour2017} as to ShallowCaps, to distinguish it from the DeepCaps~\cite{deepcaps2019} architecture} with the AlexNet~\cite{krizhevsky2012} and the LeNet~\cite{lecun1998}. For these networks, we analyze their respective memory requirements and the number of multiply-and-accumulate operations (MACs) necessary to compute an inference pass. Fig. \ref{fig:mac_memory} shows the results for the memory requirement (left) and the ratio between the MACs and the memory requirement (right). The latter is used as a comparative measure for the computational complexity. We noticed that the AlexNet has a larger memory requirement than the ShallowCaps, but with a lower MACs/Memory ratio. Hence, as shown, the ShallowCaps is more compute-intensive not only when compared to a simpler and smaller CNN like the LeNet, but also when compared to a deeper and heavier CNN like the AlexNet. This is attributed to the larger dimension of the constituent elements of the CapsNets and the high computational effort required to dynamically route the capsules.

\begin{figure}[h]
    \centering
    \vspace*{-4.3mm}
    \includegraphics[width=0.9\linewidth]{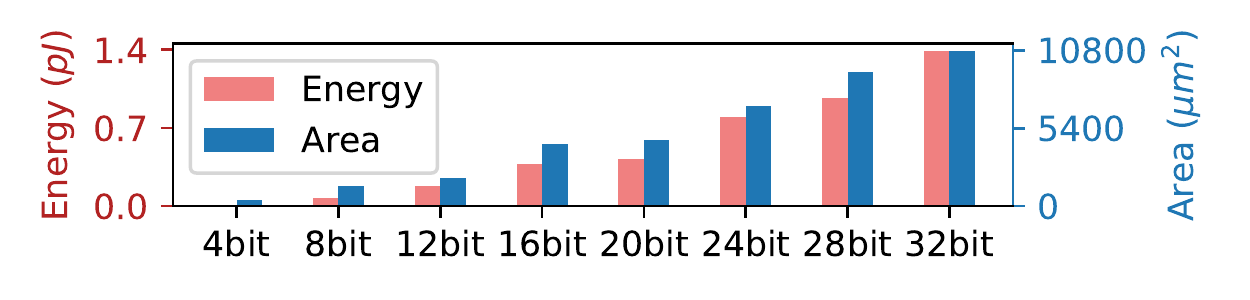}
    \vspace*{-4mm}
    \caption{\textit{Energy consumption and area footprint for a fixed-point Multiply-and-Accumulate unit (MAC) with different wordlengths.}}
    \label{fig:energy_fixedpoint}
    \vspace*{-6mm}
\end{figure}

\begin{figure}[h]
    \centering
    \vspace*{-2mm}
    \begin{minipage}[t]{.49\linewidth}
    \includegraphics[width=\linewidth]{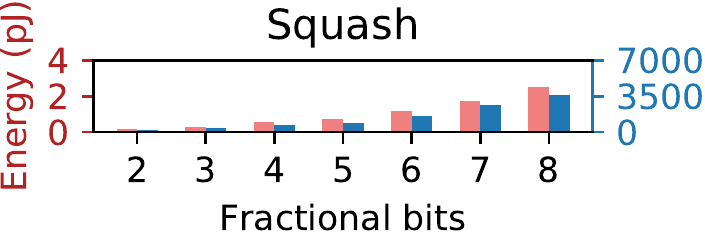}
    \end{minipage}
    \hfill
    \begin{minipage}[t]{.49\linewidth}
    \includegraphics[width=\linewidth]{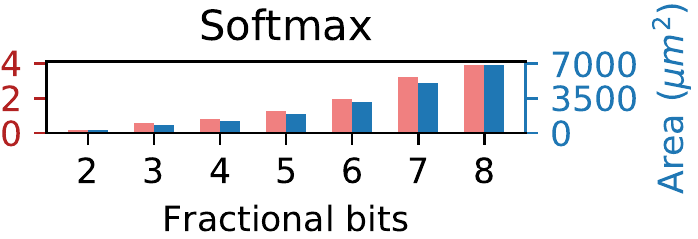}
    \end{minipage}
    \vspace*{-6mm}
    \caption{\textit{Energy consumption and area footprint for fixed-point modules performing (\textbf{left}) the squash and (\textbf{right}) the softmax with different wordlengths.}}
    \label{fig:squashhist}
    \vspace*{-2mm}
\end{figure}

\textbf{Motivational Analysis for our Target Research Problem: } Our overarching goal is to make CapsNets deployable at the edge, abandoning floating-point representation and adopting a lighter fixed-point representation. A reduction of the wordlength of the weights and activations of a CapsNet for computing the inference not only lightens the memory storage requirements, but might also have a significant impact on the energy consumption of the computational units. We perform a detailed analysis of the energy consumption and area footprint of a MAC unit, which is the basic block of specialized CapsNet accelerators like~\cite{Marchisio2019CapsAcc}, and of hardware blocks which perform computationally complex operations, i.e., \textit{squash} and \textit{softmax}, which are required during the CapsNets inference. We design different versions of a MAC unit, a squash module, and a softmax module, varying their wordlength, and we synthesize them in a UMC 65nm CMOS technology with the Synopsys Design Compiler tool to measure their area and energy consumptions. Fig. \ref{fig:energy_fixedpoint} shows that \textit{the area and energy consumption of MAC units decrease quadratically w.r.t. the wordlength}. Such analysis motivates us to focus on minimizing the wordlength to reduce the energy consumption. The results shown in Fig.~\ref{fig:squashhist} are obtained varying the number of fractional bits and keeping a single bit for the integer part. As expected, \textit{the squash and the softmax functions require more energy and area than a simple MAC operation}. The dependence of the energy consumption and of the area footprint is related quadratically to the number of fractional bits. This further motivates us to reduce the number of bits employed to perform the operations in the various layers of the CapsNets architectures.

\begin{figure}[h]
    \centering
    \vspace*{-3mm}
    \includegraphics[width=0.8\linewidth]{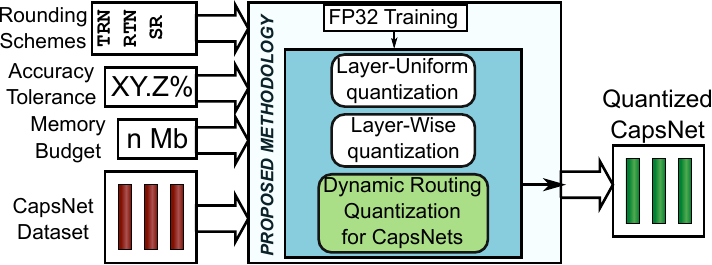}
    \vspace*{-1mm}
    \caption{\textit{An overview of our quantization framework.}}
    \label{fig:framework}
    \vspace*{-3mm}
\end{figure}

\textbf{Associated Research Challenges:} Having a too short wordlength implies lowering the accuracy of the CapsNets, which is typically an undesired outcome from the end-user perspective. To find an efficient trade-off between the memory footprint, the energy consumption and the classification accuracy, we propose a novel framework Q-CapsNets (see Fig. \ref{fig:framework}), which explores different layer-wise and operation-wise arithmetic precisions for obtaining the quantized version of a given CapsNet, with a maximum accuracy tolerance and a memory budget specified as constraints to the framework. Our approach tackles in particular the dynamic routing, which is a peculiar feature of the CapsNets and, as demonstrated in the previous paragraphs, involves complex and computationally expensive operations performed iteratively, with a significant impact on the energy consumption. 

\textbf{In a nutshell, our novel contributions are:}
\begin{itemize}[leftmargin=*]
    \item We propose a specialized framework for systematically quantizing CapsNets, given a certain accuracy tolerance (w.r.t. the full-precision CapsNet) and a certain memory budget for storing the weights. (\textbf{Section \ref{sec:framework}})
    \item Since an expensive part of CapsNets is the dynamic routing process, we further specialize the search of the numerical precision for the operations of the dynamic routing. \textit{A key advantage of using our framework}, compared to traditional DNN quantization methods, is that, as we will demonstrate in our experiments, \textit{the number of bits to route capsules can be further reduced} compared to the activations of the other layers. (\textbf{Section \ref{sec:framework}, Step 4A})
    \item We test our framework on the CapsNet model~\cite{sabour2017} on the MNIST~\cite{mnist} and Fashion-MNIST~\cite{FashionMNIST} datasets, and on the DeepCaps model~\cite{deepcaps2019} on the MNIST, FashionMNIST and CIFAR10~\cite{cifar10} datasets\footnote{To the best of our knowledge, they are the best available CapsNet models, and there is no related work able to train CapsNet models on the ImageNet~\cite{Deng2009Imagenet} dataset.}. As a key result for the latter dataset, we reduce the memory footprint by 6.2$\times$ with an accuracy loss of 0.15\%. 
    (\textbf{Section \ref{sec:results}})
    \item \textbf{Open-Source Contribution:} for reproducible research, we will release the complete source code of our framework, including the quantized CapsNet models, at https://git.io/JvDIF (Aug. 2020).
\end{itemize}

In the following \textbf{Section \ref{sec:background}}, we first discuss the CapsNets and the rounding schemes, to a level of details that is necessary to understand the rest of the paper.

\section{Background and Related Work}
\label{sec:background}
\subsection{Capsules Networks}
\label{sabour2017}
CapsNets were introduced by Hinton et al.~\cite{hinton2011}. A capsule is a group of neurons that are organized in the form of a vector, where its length (i.e., the Euclidean Norm) is the instantiation probability of a certain feature, while the individual elements of the vector encode different spatial information, like width, skew, and rotation. The main advantage of capsules is that they preserve spatial information of detected features, an important quality when performing different recognition tasks.

\begin{figure}[h]
    \centering
    \vspace*{-3mm}
    \includegraphics[width=\linewidth]{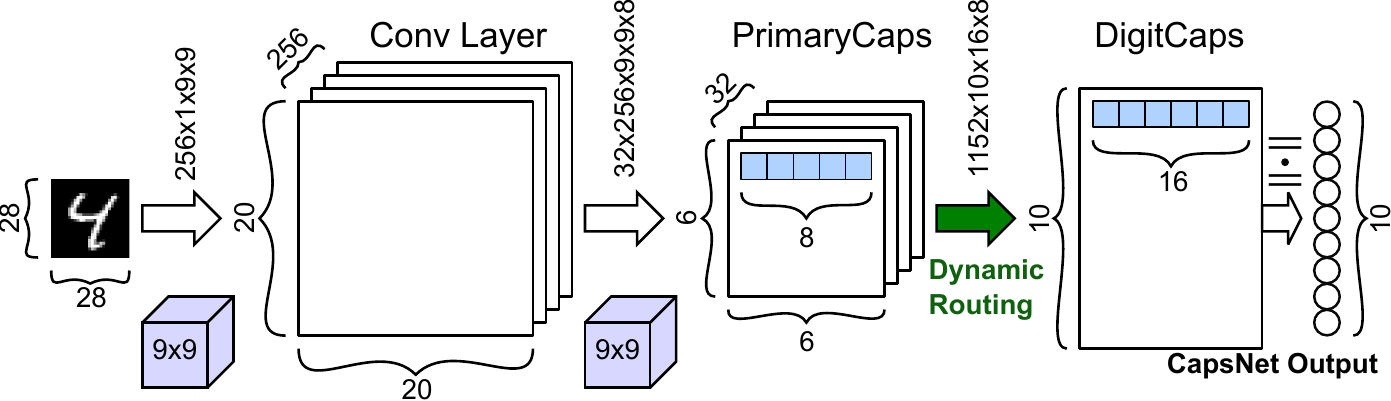}
    \vspace*{-5mm}
    \caption{\textit{CapsNet architecture for MNIST/Fashion-MNIST dataset.}}
    \label{fig:capsnetarc}
    \vspace*{-3mm}
\end{figure}
\begin{figure}[h]
	\centering
	\vspace*{-2mm}
	\includegraphics[width=\linewidth]{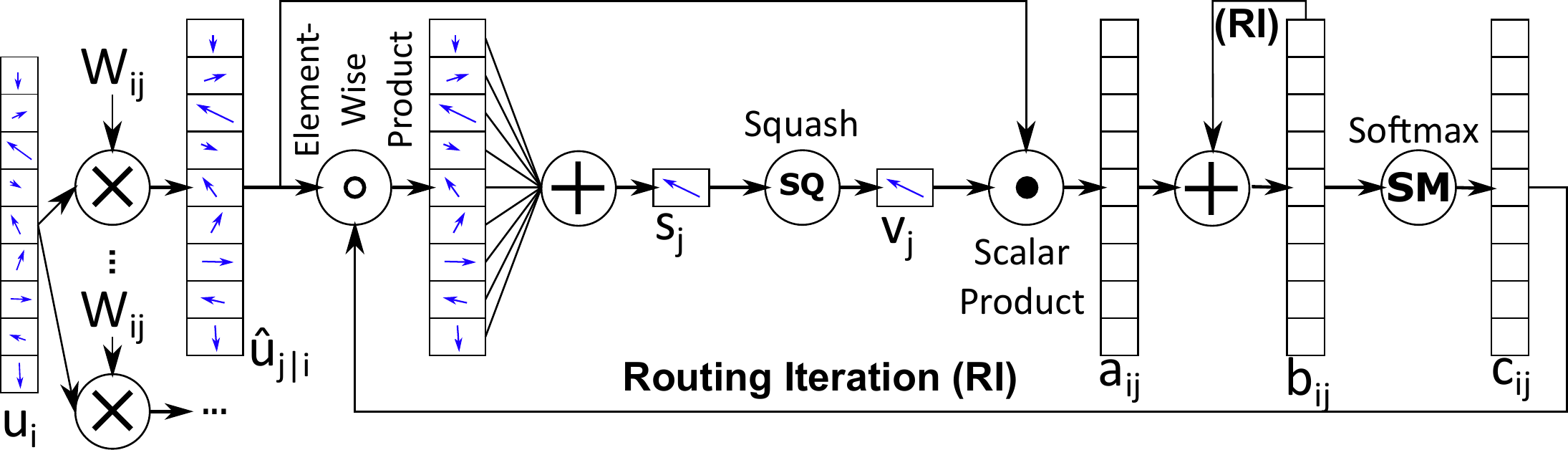}
	\vspace*{-5mm}
	\caption{\textit{The operations to be computed for the dynamic routing.}}
	\label{fig:dynamic_routing}
	\vspace*{-3mm}
\end{figure}

The architecture\footnote{Since we focus on the CapsNet inference, we do not discuss the layers and the algorithms that are {\em only} involved in the training process (e.g., decoder and reconstruction loss).} of the CapsNet proposed by Google~\cite{sabour2017} is reported in Fig. \ref{fig:capsnetarc}. It is composed of the following three layers: 
\begin{enumerate}[leftmargin=*]
    \item \textbf{(L1) Conv Layer:} 9x9 convolutional with 256 output channels; 
    \item \textbf{(L2) PrimaryCaps:} convolutional with 256 output channels. These channels are divided into 32 8-dimensional (8-D) capsules (32 8-D vectors of neurons). The \textit{squash} nonlinear function forces the length of the capsule's vector to be in the range of [0:1].
    \item \textbf{(L3) DigitCaps:} fully-connected with 16-D capsules. The number of capsules depends on the number of classes of the dataset (e.g., 10 for MNIST and FashionMNIST). Between PrimaryCaps and DigitCaps, the so-called \textit{dynamic routing} algorithm is used, as shown in Fig. \ref{fig:dynamic_routing}. 
\end{enumerate}

\begin{figure}[h]
    \centering
    \vspace*{-2mm}
    \includegraphics[width=\linewidth]{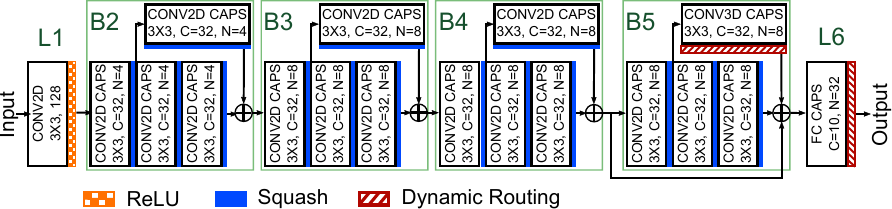}
    \vspace*{-5mm}
    \caption{\textit{DeepCaps architecture as in~\cite{deepcaps2019}.}}
    \label{fig:deepcaps}
    \vspace*{-3mm}
\end{figure}

Recently, a novel deep CapsNet architecture, \textit{DeepCaps}~\cite{deepcaps2019}, has been proposed (see Fig.~\ref{fig:deepcaps}). It introduces Convolutional layers of capsules (\textit{ConvCaps}). After the first convolutional layer with ReLU activation function, the network features 12 ConvCaps layers. Every three sequential ConvCaps layers have an additional ConvCaps layer that operates in parallel. The last parallel ConvCaps layer performs dynamic routing, while the other ConvCaps layers perform the squash function. The output layer of the DeepCaps architecture is a fully-connected capsule layer with dynamic routing. 

The dynamic routing (see Fig. \ref{fig:dynamic_routing}) is an iterative algorithm that measures the agreement between capsules in a lower layer. Each capsule is assigned to a routing coefficient. If many capsules point in the same direction with high intensity (length), they all get a high coefficient. Hence, a capsule $j$ in a higher layer is connected to all the capsules $i$ in the lower layer that mostly agree with each other. The computations are the following: 
\begin{enumerate}[leftmargin=*]
    \item Votes $\hat{u}_{j|i} = W_{ij} \times u_i$
    \item Logits initialization $b_{ij} = 0$
    \item Coupling coefficients \inlineequation[eq:softmax]{\hfill c_{ij} = \mbox{softmax}(b_{ij}) = \frac{e^{b_{ij}}}{\sum_k e^{b_{ik}}}}
    
    
    \item Preactivation $s_j = \sum_{i} c_{ij} \hat{u}_{j|i}$
    \item Activation \inlineequation[eq:squash]{\hfill v_j = \mbox{squash}(s_j) = \frac{||s_j||^2}{1 + ||s_j||^2} \frac{s_j}{||s_j||}}
    
    
    \item Agreement $a_{ij} = v_j \cdot \hat{u}_{j|i}$
    \item Logits update $b_{ij} = b_{ij} + a_{ij}$
\end{enumerate}
The dynamic routing consists of iterating the steps 3-7 for a defined number of times (e.g., 3 iterations in~\cite{sabour2017}). \textit{From a hardware perspective, such iterative computations are challenging, because they are difficult to be parallelized at a large scale.}

\subsection{Rounding Schemes}

A fixed-point number~\cite{fixedpoint} has an integer part $QI$ and a fractional part $QF$, and thus can be written as $\langle QI.QF \rangle$. The total number of bits, i.e., the wordlength $N$, is computed as the sum $NI + NF$, where $NI$ and $NF$ are the bits of the integer part and the fractional part, respectively. The precision of a fixed-point representation is $\epsilon = 2^{-NF}$, and its corresponding range of representable numbers, in a two's complement format, is [$-2^{NI-1}$, $2^{NI-1}-2^{-NF}$]. 

The \textit{rounding operation} converts a floating-point or a large-sized fixed-point number into a ``fixed-point number with shorter wordlength''. Next, we discuss the most common rounding schemes.


\textbf{Truncation (TRN)} simply removes all the extra digits from the fractional part, i.e., $x_q = \lfloor x \rfloor$.  
If we assume uniformly distributed numbers, the truncation introduces a negative average error (bias), where such error is defined as $x_q - x$.

\textbf{Round-to-Nearest (RTN)} sets a rule for approximating those values which fall exactly half-way between the two representable numbers. In particular, rounding half-up consists of rounding up these values. Considering uniformly distributed numbers, rounding-up half-way values introduces a negative average error, which is lower than the one introduced by a simple truncation.

\vspace*{-3mm}
\begin{equation}
    x_q = \lfloor x + \frac{\epsilon}{2}\rfloor
\end{equation}
\vspace*{-4mm}

\textbf{Stochastic Rounding (SR)} is defined as: 
\vspace*{-1mm}
\begin{equation}
\begin{cases}
  \lfloor x \rfloor & \mbox{if } P \geq \frac{x-\lfloor x \rfloor}{\epsilon} \\
  \lfloor x + \frac{\epsilon}{2} \rfloor & \mbox{if } P < \frac{x-\lfloor x \rfloor}{\epsilon}
\end{cases}
\end{equation}
\vspace*{-2mm}

Here, $P \in [0,1)$ is a random number with uniform distribution. The SR is an unbiased rounding scheme, but it is the most demanding one from the hardware perspective because its implementation requires 
the generation of random numbers.




\subsection{Quantization of Traditional DNNs}
Given the memory and computational requirements of DNNs, model compression is a widely studied subject where various techniques have been proposed. Han et. al~\cite{han2016} proposed Deep Compression, a three-stage pipeline to compress DNN models that combines pruning, quantization and Huffman coding, thus achieving outstanding memory reduction for different architectures. 

Focusing only on quantization, \citeauthor{courbariaux2015}~\cite{courbariaux2015} introduced BinaryConnect, constraining all the weights of a network to the two values $\{$-1, +1$\}$, while \citeauthor{hubara2016}~\cite{hubara2016} binarized both the weights and the activations. Both approaches required to train the network with binary weights. \citeauthor{gysel2016}~\cite{gysel2016} proposed the Ristretto framework, where the weights and the activations of DNN models are quantized using fixed-points, starting from a model trained in full-precision. The required numerical resolution is found with a statistical analysis of the parameters and the model is fine-tuned by retraining after the quantization. Similarly, \citeauthor{Lin2016}~\cite{Lin2016} determined the fixed-point format of the weights and activations collecting the statistics of the data and minimizing the signal-to-quantization-noise-ratio (SQNR). 

Targeting the development of efficient hardware accelerators for DNN inference, the works in~\cite{Vanhoucke2011} and~\cite{Jacob2017} tested the effect of 8-bits fixed-point quantization of the weights and the activations of different architectures, obtaining significant speed-ups at the cost of low or no accuracy reduction. Contrarily to~\cite{Vanhoucke2011} and~\cite{Jacob2017}, the works in~\cite{anwar2015} and~\cite{sakr2019} proposed a layer-wise optimization of the fixed-point representation adopted for the weights and the activations of each layer of the network. The work in~\cite{sakr2019} demonstrated that the precision required by the weights lowers for layers closer to the output, while the precision required by the activations is more constant across the layers of the network.

In our work, we introduce a novel method for quantizing the CapsNets architectures in a layer-wise fashion, tackling specifically the dynamic routing, which is peculiar for these networks. Moreover, we do not restrict the space to a single rounding scheme or to a particular data domain (weights or activation); rather our framework chooses an efficient solution to quantize different layers in a hybrid manner, thereby providing better trade-offs between the model complexity and the resulting accuracy loss.

\section{Our Q-CapsNet framework}
\label{sec:framework}

Our framework is able to progressively reduce the numerical precision of the data (e.g., weights and activations) in the CapsNet inference. During the first stage, we start with adapting/customizing the techniques for CapsNets, which are also applicable to traditional DNNs. Afterwards, we employ a specialized technique for CapsNets, which is tailored for the loops of the dynamic routing. The inputs of our framework are: 
\begin{itemize}[leftmargin=*]
    \item \textit{A CapsNet architecture}, together with the training and test dataset, and its associated architecture-specific hyperparameters.
    \item \textit{A library of rounding schemes} to choose from when quantizing the data, with the option of adopting a single rounding scheme, based on the application demand. In the first case, the framework is free to choose any rounding scheme from the library. Otherwise, it is fixed. The process of selecting an appropriate rounding scheme will be discussed in Sec. \ref{subsec:rounding_approach}.
    \item As will be explained in Sec. \ref{proposedframwork}, lowering numerical precision reduces the accuracy reached by the model. Therefore, a tolerance $acc_{TOL}$ on the loss of accuracy must be set to have a margin for quantizing the network. The target accuracy $acc_{target}$ is computed in Equation \ref{eq:acc}.
    
    \vspace*{-4mm}
    \begin{equation}
        acc_{target} = acc_{FP32} \cdot (1-acc_{TOL})
        \label{eq:acc}
    \end{equation}
    \vspace*{-5mm}
    
    \item \textit{Maximum memory budget} that can be occupied for the storage of the quantized weights and biases.
\end{itemize}

Our Q-CapsNet framework aims at satisfying both requirements on accuracy and memory usage. An effective way to reduce the model's memory usage is through aggressively quantizing the weights. We perform this operation in the steps (1) and (2) of the proposed framework. Once the memory budget is satisfied, if there is still some margin on the tolerable accuracy loss, we reduce the numerical precision of the weights and activations, to reduce the energy consumed during the CapsNet inference computations, and the framework returns the \verb|model_satisfied|. Otherwise, if a solution which satisfies both the requirements on the accuracy and the memory usage cannot be found, our framework returns two sub-optimal solutions as the followup:

\begin{enumerate}[leftmargin=*]
    \item[I] \verb|model_accuracy|: A quantized CapsNet with the target accuracy and the minimum possible memory footprint (which can be slightly higher than the budget); 
    \item[II] \verb|model_memory|: A quantized CapsNet that satisfies the memory requirements, and achieving the maximum possible accuracy (which can be slightly lower than the target). 
\end{enumerate}

\subsection{Step-by-Step Description of our Framework}
\label{proposedframwork}
As a preliminary stage, a given input CapsNet is trained in full-precision (32-bits floating-point), whose accuracy is denoted as $acc_{FP32}$. From $acc_{FP32}$ and the accuracy tolerance ($acc_{TOL}$, input of the framework), we compute the target accuracy ($acc_{target}$) as in Equation \ref{eq:acc}. The procedure followed for quantizing the given CapsNet (see Figure \ref{fig:methodology} and Algorithm \ref{alg:methodology}) is composed of the following steps:

\begin{figure}[h]
    \centering
    \vspace*{-2mm}
    \includegraphics[width=0.95\linewidth]{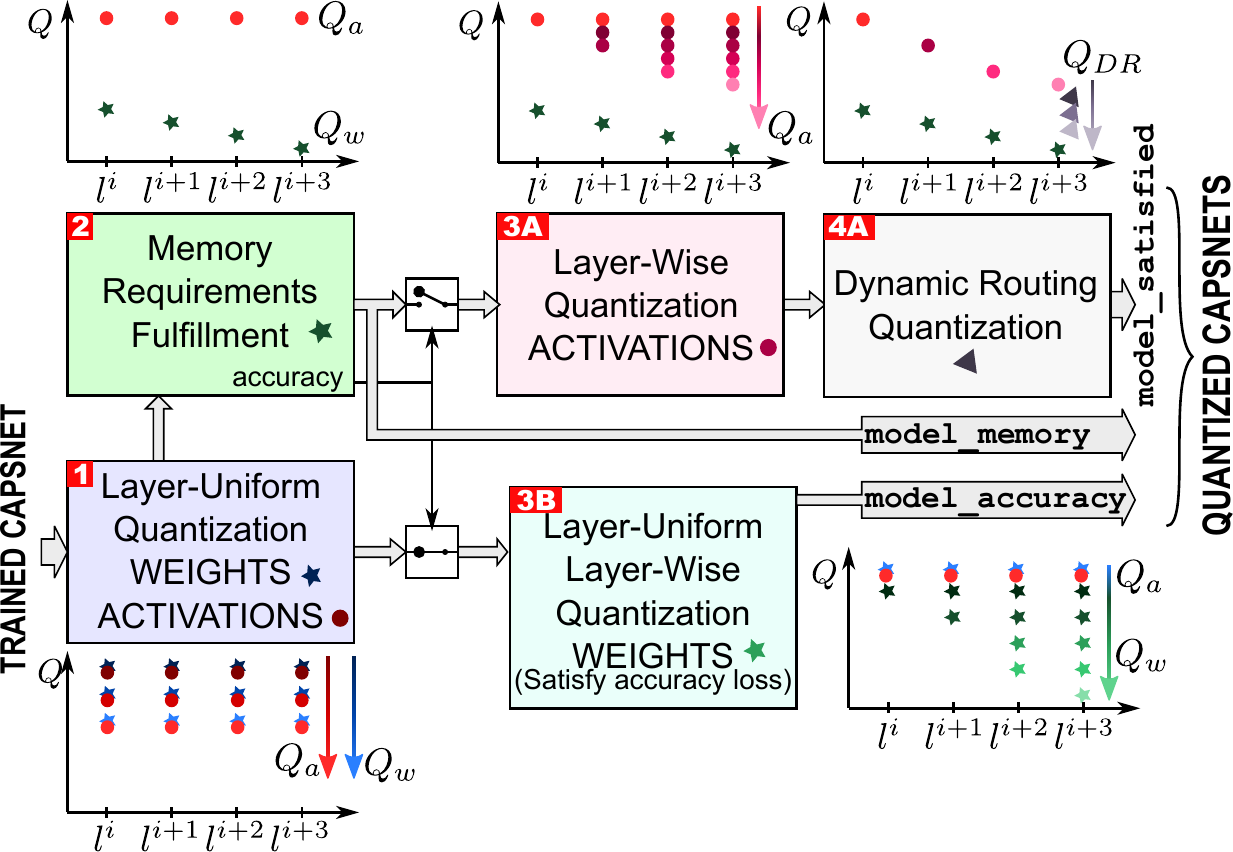}
    \vspace*{-1.5mm}
    \caption{\textit{Flow of Our Framework for Quantizing CapsNets.}}
    \label{fig:methodology}
    \vspace*{-3mm}
\end{figure}

\begin{figure}[h]
\vspace*{-5mm}
 \begin{algorithm}[H]
 \captionsetup{font=footnotesize}
  \caption{Pseudo-Code of Our Framework. (see Fig. \ref{fig:methodology} for its Flow)}
  \label{alg:methodology}
  \vspace*{-1mm}
  \begin{small}
 \begin{algorithmic}[1]
 
  \Procedure {Q-CapsNet}{CapsNet, $acc_{TOL}$, memory\_budget}
  
  \LineComment{Full Precision training}
  \State {model, $acc_{FP32} \gets$ train(CapsNet)}
  \State {$acc_{target} = acc_{FP32}(1-acc_{TOL})$}
  
  \LineComment {\textit{Step 1)}}
  \State {$acc_{step1} = acc_{FP32}(1-acc_{TOL}\cdot 0.05)$}
  \State{model, Q $\gets$ BinarySearch(model, (weights, act), $Q_{init}=32$, $acc_{min}=acc_{step1}$)}
  \State {$(Q_{w,s1})_l = Q; \hspace{2mm} (Q_{a,s1})_l = Q$ $ \forall l$}
 
  \LineComment {\textit{Step 2)}}
  \State {[$(Q_{w,mm})_0$, ..., $(Q_{w,mm})_L$] $\gets$ Eq.\ref{eq:weights}(params $P$, memory\_budget)}
  \State{model\_memory, $acc_{mm} \gets$ test(quant(model, weights $\gets Q_{w,mm}$, act $\gets Q_{a,s1}$ ))}
  
  \If{$acc_{mm} > acc_{target}$}
    \LineComment{\textit{Step 3A)}}
    \State{model, $Q_a \gets$ LayerWise(model, act, $Q_{init}=Q_{a,s1}$, $acc_{min} = acc_{target} + 0.5(acc_{mm}-acc_{target})$)}
    \LineComment{\textit{Step4A)}}
    \For {each layer $l$ with dynamic routing}
    \State {model, $(Q_a)_l \gets$ DRquant(model, model.DRact$_l$, $Q_{init}=(Q_a)_l$, $acc_{min}=acc_{target}$)}
    \EndFor
    \State{\textbf{end for}}
    \State{\textbf{return} model\_satisfied}
    
 \Else 
 \LineComment{\textit{Step3B)}}
 \State{model, $Q_w \gets$ BinarySearch(model, weights, $Q_{init}=Q_{w,1}$, $acc_{min} = acc_{target}$)}
 \State{model\_accuracy, $Q_w \gets$ LayerWise(model, weights, $Q_{init}=Q_w$, $acc_{min}=acc_{target}$}
 \State{\textbf{return} model\_memory, model\_accuracy}
 \EndIf
 \State{\textbf{end if}}
 \EndProcedure
 \end{algorithmic}
 \end{small} 
 \vspace*{0mm}
 \end{algorithm}
 \vspace*{-6mm}
 \end{figure}

\begin{enumerate}[leftmargin=*,align=left, wide=0pt]
    \item[1)] \textbf{Layer-Uniform Quantization (weights + activations)}: 
    We convert all weights and activations to a fixed-point arithmetic, with 1-bit integer part, and $Q_w$-bit and $Q_a$-bit fractional part, respectively. Afterwards, we further reduce their precision in a uniform way (e.g., $Q_w=Q_a$). In this stage, only 5\% of the $acc_{TOL}$ is consumed. To find the correct wordlength of $Q_w$ and $Q_a$, we use a binary search algorithm \cite{binarysearch}.

    \item[2)] \textbf{Memory Requirements Fulfillment}: In this stage, we quantize only the CapsNet weights. Following the idea of Raghu et al. \cite{raghu2017a} that perturbations to weights in final layers can be more costly than perturbations in the earlier layers, we set for each layer $l$ its respective $Q_w$ such that $(Q_w)_{l+1}=(Q_w)_{l}-1$. Having set these conditions, we can compute the correct $Q_w$ as the maximum integer value that satisfies the Equation \ref{eq:weights}, where $L$ is the total number of layers, $M$ is the memory budget, and $P^l$ is the number of parameter (weights) in the layer $l$.
    
    \vspace*{-5mm}
    \begin{equation}
        \sum_{l = 0}^{L-1} \left( P^l \cdot \left( (Q_w)_0 -l\right) \right) \leq M
        \label{eq:weights}
    \end{equation}
    \vspace*{-3mm}
    
    With this rule, we obtain a quantized CapsNet model, denoted as \verb|model_memory|, which fulfills the memory requirements. Afterwards, we test the accuracy of the \verb|model_memory|, denoted as $acc_{mm}$ and compare it to $acc_{target}$. Based on its results, the next step can take two directions. If $acc_{mm}$ is higher, we continue to (3A) for further quantization steps. Otherwise, it jumps to (3B).

    
    
    \item[3A)] \textbf{Layer-Wise quantization of activations}: To quantize the activations, we start from the initial $Q_a$, as computed during the step (1). As shown in Algorithm \ref{alg:layerwise}, we proceed in a layer-wise fashion. During the first step, each layer of the CapsNet (except the first one) is selected, and $Q_a$ is lowered until the minimum value for which the accuracy remains higher than $acc_{target}$. Afterwards, the wordlength of the first two layers is fixed, while we further reduce $Q_a$ for all but the first layers. We repeat this step iteratively until the $Q_a$ for the last layer is set.
    
\begin{figure}[h]
\vspace*{-6mm}
 \begin{algorithm}[H]
 \captionsetup{font=footnotesize}
  \caption{Algorithm for Layer-wise Quantization}
  \label{alg:layerwise}
  \vspace*{-1mm}
  \begin{small}
 \begin{algorithmic}[1]
 \State Given: $Q_{init}$ initial number of quantization bits to start the algorithm, $acc_{min}$ minimum value of accuracy that can be reached.
  \Procedure {LayerWise}{model, params, $Q_{init}$, $acc_{min}$}
  \State {$Q$ = [$(Q)_0$, $(Q)_1$, ..., $(Q)_L$], $(Q)_l = Q_{init}$} 
  \State {$StartL = 1$}
  \While {$StartL < L$}
     \State {$acc = 100$} 
         \While {$acc \geq acc_{min}$}
             \State {$(Q)_l \gets (Q)_l-1$, $l \in [StartL, ..., L]$}
             \State {model, $acc$ = test(quant(model, params $\gets Q$))}
         \EndWhile 
         \State{\textbf{end while}}
     \State {$(Q)_l \gets (Q)_l+1$, $l \in [StartL, ..., L]$}
     \State {$StartL \gets StartL +1$}
  \EndWhile 
  \State{\textbf{end while}}
  \State \Return {quant(model, params $\gets Q$), $Q$}
  \EndProcedure
 \end{algorithmic}
 \end{small}
 \vspace*{0mm}
 \end{algorithm}
 \vspace*{-5mm}
 \end{figure}

    
    \item[4A)] \textbf{Dynamic Routing Quantization}: The dynamic routing is computationally expensive due to the complex operations, such as \textit{squash} (Eq. \ref{eq:squash}) and \textit{softmax} (Eq. \ref{eq:softmax}), and the operations are performed iteratively. Hence, the wordlength of its arrays may be different as compared to other layers of the CapsNet. This step operates only on the data involving the \textit{squash} and \textit{softmax} operations. A specialized quantization process is performed in this step, as shown in Fig. \ref{fig:dynr_quant} and Algorithm \ref{alg:dynr}. As we will demonstrate in our experiments, the operators of the dynamic routing can be quantized more than the other activations (i.e., with a wordlength lower than $Q_a$, which we call $Q_{DR}$). The quantized CapsNet model that is generated at the end of this step is denoted as \verb|model_satisfied|.

    
    \begin{figure}[h]
        \centering
        \vspace*{-2mm}
        \includegraphics[width=\linewidth]{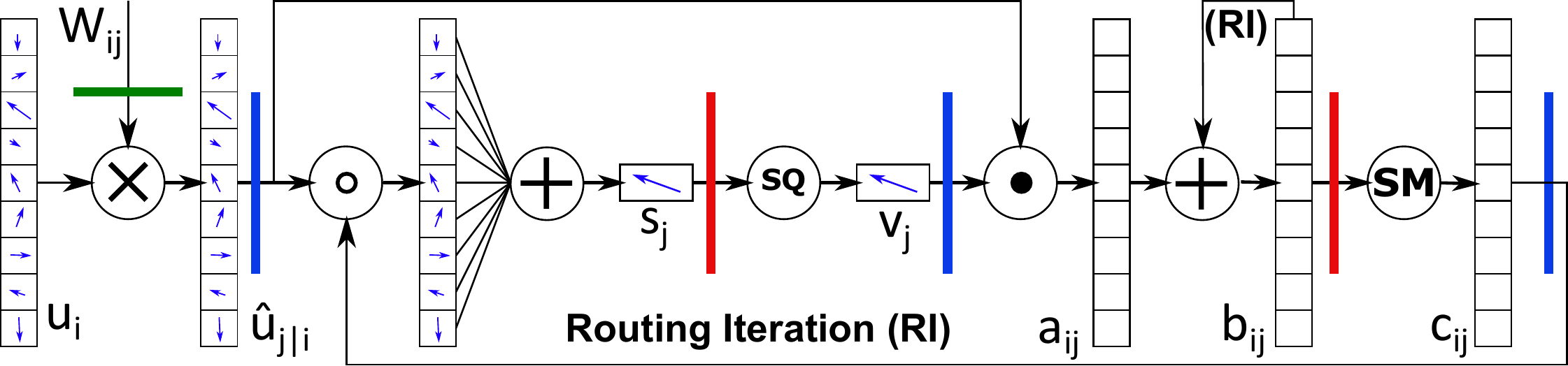}
        \vspace*{-5mm}
        \caption{\textit{Quantization of a capsule layer with dynamic routing. Colored bars show the arrays that are rounded and quantized. In green, the weights are quantized with $Q_w$ bits. In blue, the activations are quantized with $Q_a$ bits. In red, data are quantized more aggressively with $Q_{DR}$ bits. The precision is lowered before complex and compute-intensive functions (squash, softmax).}}
        \label{fig:dynr_quant}
        \vspace*{-3mm}
    \end{figure}

 \begin{figure}[h]
 \vspace*{-2mm}
 \begin{algorithm}[H]
 \captionsetup{font=footnotesize}
  \caption{Algorithm for Dynamic Routing Quantization}
  \label{alg:dynr}
  \vspace*{-1mm}
 \begin{small}
 \begin{algorithmic}[1]
 \State Given: $Q_{init}$ initial number of quantization bits to start the algorithm, $acc_{min}$ minimum value of accuracy that can be reached.
 \Procedure {DRquant}{model, params, $Q_{init}$, $acc_{min}$}
  \State {$Q = Q_{init}$} 
  \State {$acc=100$}
  \While {$acc \geq acc_{min}$}
        \State {$Q \gets Q-1$}
        \State {model, $acc$ = test(quant(model, params $\gets Q$))}
    \EndWhile 
    \State{\textbf{end while}}
     \State {$Q \gets Q+1$}
  \State \Return {quant(model, params $\gets Q$), $Q$}
  \EndProcedure
 \end{algorithmic}
 \end{small}
 \vspace*{0mm}
 \end{algorithm}
 \vspace*{-8mm}
 \end{figure}
    
    \item[3B)] \textbf{Layer-Uniform ad Layer-Wise Quantization of Weights}: Starting from the outcome of step (1), we quantize the weights only, first in a uniform and then in a layer-wise manner (as in step 3A) until reaching $acc_{target}$. The resulting CapsNet model (\verb|model_accuracy|) is returned as the output of the framework, together with \verb|model_memory|, as generated in step (2).
    
\end{enumerate}
 
 \subsection{Rounding Scheme Selection}
 \label{subsec:rounding_approach}
 
 For each rounding scheme from the given library, its corresponding quantized model is generated. Hence, \textit{our framework executes the Algorithm \ref{alg:methodology} for each rounding scheme in parallel}. Note, due to different rounding errors, it is possible that for one rounding scheme our framework executes the Path A, while for another schemes it executes the Path B. At the end of the execution of all branches, the best rounding scheme within the library is selected with the following criteria, depending on whether the algorithm has followed Path A or not.
\begin{enumerate}[leftmargin=*,align=left, wide=0pt]
    \item[A)] \textbf{There are some models generated from Path A: }
    \begin{enumerate}[leftmargin=*,align=left, wide=0pt]
        \item[1)] Models from Path B are discarded.
        \item[2)] The model with lower memory is selected.
        \item[3)] With the same memory, the model with fewer bits used to represent activations is selected.
        \item[4)] With the same memory and bits for the activations, the model with the simplest rounding scheme is selected, e.g., with our examples, in order, truncation, round-to-nearest-even, and stochastic rounding. Note, while the first one simply requires the deletion of the LSBs, the last one requires more complex operations to decide the orientation of the rounding. 
    \end{enumerate}
    \item[B)] \textbf{There are models only from Path B: }
    \begin{enumerate}[leftmargin=*,align=left, wide=0pt]
        \item[1)] In this case, two models are returned. Selecting from \verb|memory_model|, the model with the highest-possible accuracy is returned. 
        \item[2)] Selecting from \verb|accuracy_model|,  the model with the lowest-possible memory is returned. 
        \item[3)] If more than one model have the same highest accuracy and the lowest memory, the simplest rounding scheme is preferred to break the tie. 
    \end{enumerate}
\end{enumerate}

\vspace*{-3mm}
\section{Results}
\label{sec:results}
\subsection{Experimental setup}

\begin{figure}[h]
    \centering
    \vspace*{-2mm}
    \includegraphics[width=0.9\linewidth]{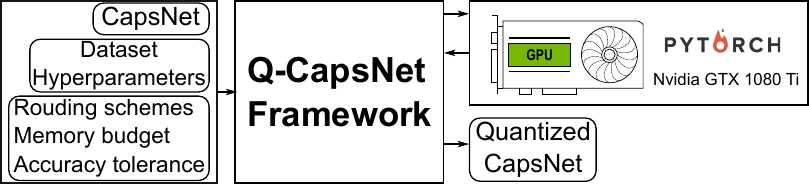}
    \caption{\textit{Experimental setup to test our Q-CapsNet framework.}}
    \vspace*{-1mm}
    \label{fig:exp_setup}
    \vspace*{-3mm}
\end{figure}

We implement the Q-CapsNet framework (see Fig. \ref{fig:exp_setup}) in PyTorch \cite{paszke2017}, and we run it on two Nvidia GTX 1080 Ti GPUs. We test it on the CapsNet model proposed by Google \cite{sabour2017}, i.e., ShallowCaps, also previously described in Sec. \ref{sabour2017}, for MNIST \cite{mnist} and FashionMNIST \cite{FashionMNIST} datasets, and on the DeepCaps model for the MNIST, FashionMNIST and CIFAR10 \cite{cifar10} datasets. The MNIST database is a collection of 28x28 grayscale handwritten digits, from 1 to 10, composed of 60,000 training samples and 10,000 testing samples. The FashionMNIST is a collection of 28x28 grayscale images, representing Zalando's articles associated to 10 different classes. It is composed of 60,000 training samples and 10,000 testing samples. The CIFAR10 is a collection of 32x32 color images organized in 10 different classes, with the training set composed of 50,000 samples and the testing set of 10,000 samples. 
For full precision training, data augmentation is achieved as follows: 
\begin{itemize}[leftmargin=*]
    \item MNIST: images are randomly shifted by maximum two pixels and rotated of 2 degrees; 
    \item FashionMNIST: images are randomly shifted of 2 pixels and horizontally flipped with a probability of 0.2; 
    \item CIFAR10: images are resized to 64x64\footnote{The original images of size 32x32 are resized to 64x64 by bilinear interpolation, to allow deeper networks, as reported in the original paper \cite{deepcaps2019}.}, randomly shifted of 5 pixels, rotated of 2 degrees and horizontally flipped with a probability of 0.5. 
\end{itemize}
No data augmentation is done on the images for testing.


\subsection{Quantized Architectures}

\noindent
\textbf{ShallowCaps for the MNIST Dataset} \\
The ShallowCaps architecture \cite{sabour2017} is trained in full precision (FP32) on the MNIST dataset, 
for 100 epochs and with batch size equal to 100. We use an exponential decay learning policy, with an initial learning rate of 0.001, 2000 decay steps and 0.96 decay rate. Its achieved test accuracy is 99.67\%.

Afterwards, the framework proceeds as described in Sec. \ref{proposedframwork}, with the aim of concurrently satisfying the memory and accuracy requirements. Since the algorithm has a conditional path, for the sake of clarity, we present two examples, which correspond to the execution of the different branches of the algorithm.

\begin{figure}[h]
    \centering
    \vspace*{-2mm}
    \includegraphics[width=.95\linewidth]{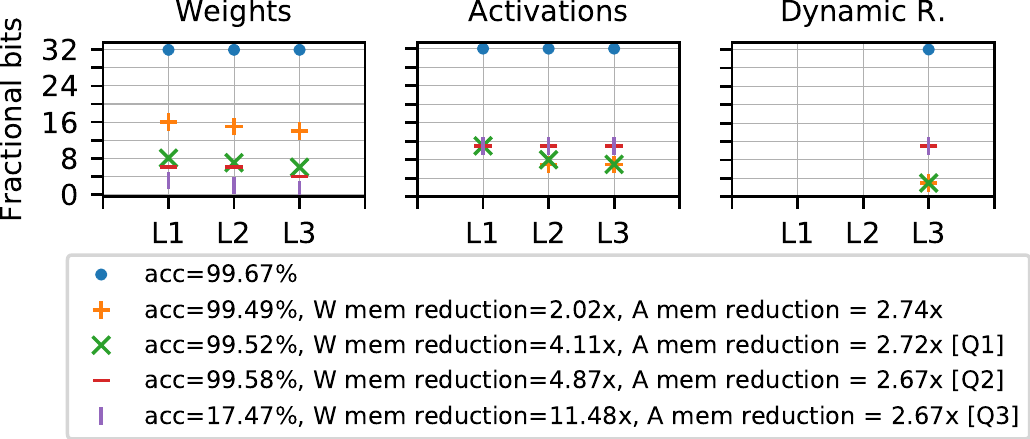}
    \vspace*{-1mm}
    \caption{\textit{Q-CapsNet results of the ShallowCaps \cite{sabour2017} for the MNIST \cite{mnist} dataset.}}
    \label{fig:capsres}
    \vspace*{-3mm}
\end{figure}

\underline{Test of the Path A:} For the first set of experiments, we test the Path A of the framework, i.e., when both the memory and accuracy constraints are satisfied. Since the memory requirement at FP32 is 217Mbit, we set the memory budget equal to 45Mbit, with an accuracy tolerance of 0.2\%. The results in Fig. \ref{fig:capsres} [Q1] show that the \verb|model_satisfied| reduces the memory footprint of the weights by 4.11$\times$, as compared to the FP32 model, with an accuracy equal to 99.52\%. Along with the reduction of the memory occupied by the weights (W mem), we report the reduction of the memory required to store the activations (A mem). For \verb|model_satisfied|, this memory footprint is reduced of 2.72$\times$. 


\underline{Test of the Path B:} Since our framework executes the Path B if it cannot find a solution which satisfies both requirements, for its testing purpose, we specify very low memory budgets as the input. The results of our experiments, shown in Fig \ref{fig:capsres}, indicate that to satisfy the memory requirements, weights of \verb|model_memory| [Q3] are set to very low wordlengths, causing an extreme reduction of accuracy. To satisfy the accuracy requirements in \verb|memory_accuracy| [Q2], weights are reduced to the minimum possible wordlength. 

\noindent
\textbf{ShallowCaps for the FashionMNIST Dataset} \\
Similar sequences of tests, with a set of memory budget and accuracy tolerance specifications, are performed on the same ShallowCaps architecture for the FashionMNIST dataset. The results from our experiments are reported in Table \ref{tab:results}. 

\begin{table}[t]
\centering
\vspace*{2mm}
    \caption{Q-CapsNet's accuracy results, weight~(W) memory and activation~(A) memory reduction for the ShallowCaps \cite{sabour2017} and for the DeepCaps \cite{deepcaps2019} on MNIST~\cite{mnist}, Fashion-MNIST~\cite{FashionMNIST} and CIFAR10~\cite{cifar10} datasets.}
    \label{tab:results}
    \vspace*{-1mm}
    \resizebox{.95\linewidth}{!}{%
    \begin{tabular}{| c | c | c | c | c |}
    \hline 
    \textbf{Model} & \textbf{Dataset} & \textbf{Accuracy} & \textbf{W mem reduction} & \textbf{A mem reduction} \\
    \hline 
    ShallowCaps & MNIST & 99.58\% & 4.87x & 2.67x \\
    \hline 
    ShallowCaps & MNIST & 99.49\% & 2.02x & 2.74x \\
    \hline 
    ShallowCaps & FMNIST & 92.76\% & 4.11x & 2.49x \\
    \hline 
    ShallowCaps & FMNIST & 78.26\% & 6.69x & 2.46x \\
    \hline 
    DeepCaps & MNIST & 99.55\% & 7.51x & 4.00x \\
    \hline 
    DeepCaps & MNIST & 99.60\% & 4.59x & 6.45x \\
    \hline 
    DeepCaps & FMNIST & 94.93\% & 6.4x & 3.20x \\
    \hline 
    DeepCaps & FMNIST & 94.92\% & 4.59x & 4.57x \\
    \hline
    DeepCaps & CIFAR10 & 91.11\% & 6.15x & 2.50x \\
    \hline
    DeepCaps & CIFAR10 & 91.18\% & 3.71x & 3.34x \\
    \hline
    \end{tabular}
    }
    \vspace*{-4mm}
\end{table}

\begin{figure}[h]
    \centering
    \vspace*{-5mm}
    \includegraphics[width=0.95\linewidth]{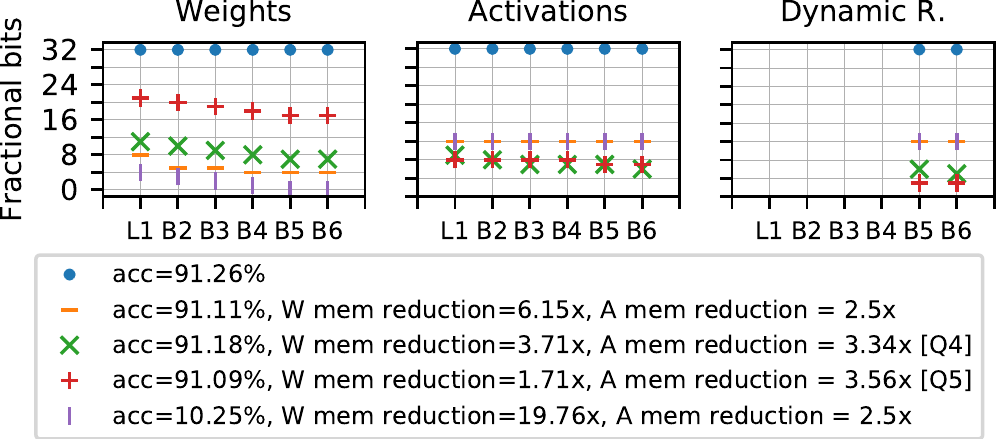}
    \caption{\textit{Q-CapsNet results of the DeepCaps \cite{deepcaps2019} for CIFAR10 \cite{cifar10} dataset.}}
    \vspace*{-1mm}
    \label{fig:deepres}
    \vspace*{-4mm}
\end{figure}

\noindent
\textbf{DeepCaps for MNIST, FashionMNIST and CIFAR10 datasets} \\
Several tests are carried out on the DeepCaps architecture. We mainly discuss the results obtained with the SR scheme, which outperforms the other (simpler) rounding schemes. The DeepCaps architecture trained in full-precision on the MNIST dataset achieves a 99.75\% accuracy, on par with the accuracy obtained in \cite{sabour2017}, while on the FashionMNIST it achieves a 95.08\% accuracy. Table \ref{tab:results} reports some key results obtained with the Q-CapsNet framework on these two datasets. Fig. \ref{fig:deepres} reports graphically some key results obtained with the Q-CapsNet framework on the DeepCaps for the CIFAR10 dataset. 

\subsection{Comparison between Different Rounding Schemes} 
Experiments performed for different inputs to the framework show that truncation and round-to-nearest schemes return identical results. This is due to the fact that these schemes differ from each other only for a very small set of continuous values, i.e., those falling half-way between two discrete values, and therefore the influence on the final results of the network is negligible. 

\begin{figure}[t]
    \centering
    \vspace*{-3mm}
    \includegraphics[width=\linewidth]{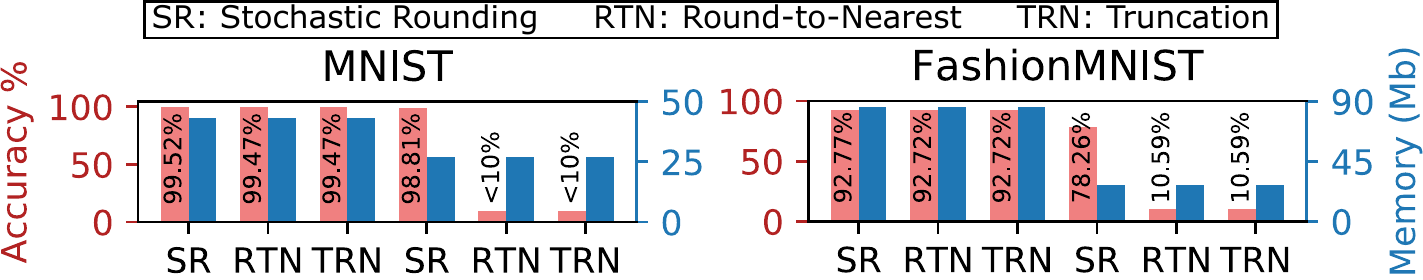}
    \vspace*{-5mm}
    \caption{\textit{Accuracy and memory comparison for Q-CapNet models of the ShallowCaps architecture, obtained using different rounding schemes. \textbf{(left)} Results for MNIST. \textbf{(right)} Results for Fashion-MNIST.}}
    \label{fig:comparison_rounding}
    \vspace*{-6mm}
\end{figure}

Fig. \ref{fig:comparison_rounding} shows the accuracy reached by the ShallowCaps when different rounding schemes are applied, with the same memory usage. For both the MNIST and FashionMNIST datasets, stochastic rounding outperforms simpler methods, e.g., when a lower memory footprint is required. 
Indeed, the stochastic rounding presents the advantage of randomizing the quantization noise. Small values close to zero have a non-null probability of being rounded up rather than always being forced to zero. This solution avoids an excessive loss of information when iteratively performing computations and quantizations.

\vspace*{-1mm}
\subsection{Further Discussion on the Results}

By considering the occupied weight memory and the accuracy as the evaluation metrics, we noticed that, usually, the \verb|model_satisfied| seems to be Pareto-dominated by the \verb|model_accuracy|, like in the case of Q1 and Q2 in Fig.~\ref{fig:capsres}, and of Q4 and Q5 in Fig.~\ref{fig:deepres}. However, since Q1 and Q5 have lower wordlengths for the activations and the dynamic routing, compared respectively, to Q2 and Q4, the potential energy-efficiency gains for its computations using MAC operators, \textit{squash} and \textit{softmax} (recall Figures~\ref{fig:energy_fixedpoint} and~\ref{fig:squashhist}) are huge, even with a small change in the activation memory. Note, the wordlength for the dynamic routing operations can be reduced up to 3 or 4 bits with very limited accuracy loss compared to the full-precision model. Such an outcome is attributed to a common feature of the dynamic routing. The operations of the involved coefficients (along with \textit{squash} and \textit{softmax}, see Fig.~\ref{fig:dynamic_routing}) are updated dynamically, thereby adapting to the quantization more easily than previous layers like Conv Layer and PrimaryCaps. Hence, these computations can tolerate a more aggressive quantization.

\vspace*{-3mm}
\section{Conclusion}
We proposed a specialized framework for quantizing CapsNets, called Q-CapsNets. We exploited the peculiar features of CapsNets, occurring during the dynamic routing, for designing a quantization methodology that enables further precision reduction of the wordlength while a certain accuracy loss is tolerated. Our Q-CapsNets framework produces compact yet accurate quantized CapsNet models. Hence, it represents the first step towards designing energy-efficient CapsNets, and could potentially open new avenues towards the large-scale adoption of CapsNets for inference in a resource-constrained scenario.

\vspace*{-1mm}
\section*{Acknowledgments}
\vspace*{-1mm}
\begin{scriptsize}
\noindent
This work has been partially supported by the Doctoral College Resilient Embedded Systems which is run jointly by TU Wien's Faculty of Informatics and FH-Technikum Wien.
\end{scriptsize}

\vspace*{-1mm}
\begin{refsize}

\end{refsize}

\end{document}